\documentclass[twoside,11pt]{article}

\usepackage{jmlr2e}
\usepackage{graphicx}
\usepackage[T1]{fontenc}
\usepackage{graphics}
\usepackage{algorithm,amsmath, amssymb}
\usepackage{algorithmic}
\usepackage{epsfig,url}
\usepackage{setspace}
\usepackage{booktabs}
\usepackage{bm,bbm}

\newcommand{\argmin}{\operatornamewithlimits{argmin}}
\newcommand{\eps}{\epsilon}
\newcommand{\E}{\mathbb{E}}

\jmlrheading{14}{2013}{xx-xx}{05/12}{05/13}{Paramveer S. Dhillon, Dean P. Foster, Sham M. Kakade and Lyle H. Ungar}

\ShortHeadings{A Risk Comparison of Ordinary Least Squares vs Ridge Regression}{Dhillon, Foster, Kakade and Ungar}
\firstpageno{1}

\begin{document}

\title{A Risk Comparison of Ordinary Least Squares vs Ridge Regression}

\author{\name Paramveer S. Dhillon \email dhillon@cis.upenn.edu \\
       \addr Department of Computer and Information Science\\
       University of Pennsylvania\\
       Philadelphia, PA 19104, USA
       \AND
       \name Dean P.  Foster \email foster@wharton.upenn.edu \\
       \addr Department of Statistics\\
       Wharton School, University of Pennsylvania\\
       Philadelphia, PA 19104, USA
        \AND
        \name Sham M.  Kakade \email skakade@microsoft.com \\
       \addr Microsoft Research\\
       One Memorial Drive\\
      Cambridge, MA 02142, USA
        \AND
       \name Lyle H. Ungar \email ungar@cis.upenn.edu \\
       \addr Department of Computer and Information Science\\
       University of Pennsylvania\\
       Philadelphia, PA 19104, USA}

\editor{Gabor Lugosi}
\maketitle

\begin{abstract}%
We compare the risk of ridge regression to a simple variant of ordinary
least squares, in which one simply projects the data onto a finite dimensional subspace
(as specified by a principal component analysis) and then performs an
ordinary (un-regularized) least squares regression in
this subspace. This note shows that the risk of this ordinary least squares
method (PCA-OLS) is within a constant factor (namely 4) of the risk of
ridge regression (RR).
\end{abstract}

\section{Introduction}

Consider the fixed design setting where we have a set of $n$
vectors  ${\mathcal{X}}= \{X_i\}$, and let {\bf X}
denote the matrix where the ${\it i^{th}} $
row of {\bf X} is ${X}_i$. The observed label vector is ${Y} \in
\mathbb{R}^n$. Suppose that:
\[
{Y} = {\bf X}\beta +\eps,
\]
where $\eps$ is independent noise in each coordinate, with the variance of
$\eps_i$ being $\sigma^2$.

The objective is to learn $ \mathbb{E}[ {Y} ]={\bf X}\beta$.  The expected loss of a vector $\beta$
estimator is:
\[
{L}(\beta)= \frac{1}{n}\mathbb{E_Y}[\|Y-{\bf X}\beta\|^2],
\]
Let $\hat\beta$ be an
estimator of $\beta$ (constructed with a sample $Y$).
Denoting 
\begin{equation}
\nonumber
{\bf \Sigma} := \frac{1}{n} {\bf X}^T{\bf X},
\end{equation}
we have that the risk (i.e., expected excess loss) is:
\begin{equation}
\nonumber
\textrm{Risk}(\hat\beta):=\E_{\hat\beta} [L(\hat{\beta})-L(\beta)] = \E_{\hat\beta}\|\hat{\beta} - \beta\|^2_{\bf \Sigma},
\end{equation}
where $\|x\|_{\bf \Sigma}=x^\top{\bf \Sigma}x $ and where the
expectation is with respect to the randomness in $Y$.

We show that a simple variant of ordinary (un-regularized) least
squares always compares favorably to ridge regression (as measured by
the risk). This observation is based on the following bias variance decomposition:
\begin{equation}\label{biasDecomp}
\mathrm{Risk} (\hat{\beta})
= \underbrace{\mathbb{E}\|\hat{\beta} - \bar{\beta}\|^2_{\bf \Sigma}}_{\mbox{Variance}}+ \underbrace{\|\bar{\beta} - \beta\|^2_{\bf \Sigma}}_{\mbox{Prediction Bias}},
\end{equation}
where $\bar{\beta}=\mathbb{E}[\hat{\beta}]$.

\subsection{The Risk of Ridge Regression (RR)}

Ridge regression or Tikhonov Regularization~\citep{ridge} penalizes the $\ell_2$ norm of a parameter vector
$\beta$ and ``shrinks'' it towards zero, penalizing
large values more.  The estimator is:
\begin{equation}
\nonumber
\hat{\beta}_{\lambda} = \argmin_{\beta}\{ \|Y - {\bf X}\beta \|^2 + \lambda\|\beta\|^2  \}.
\end{equation}
The closed form estimate is then:
\begin{equation}
\nonumber
\hat{\beta}_{\lambda} =({\bf \Sigma} + \lambda {\bf
  I})^{-1}\left(\frac{1}{n}{\bf X}^TY\right).
\end{equation}
Note that 
\[
\hat\beta_0=\hat\beta_{\lambda=0} = \argmin_{\beta}\{ \|Y - {\bf X}\beta \|^2 \},
\]
is the ordinary least squares estimator.

Without loss of generality, rotate {\bf X} such that: 
\begin{equation*}
\label{sigmaRot}
{\bf \Sigma} = \textrm{diag}(\lambda_1,\lambda_2,\ldots,\lambda_p),
 \end{equation*}
where the $\lambda_i$'s are ordered in decreasing order.

To see the nature of this shrinkage observe that:

\begin{equation*}
[\hat{\beta}_{\lambda}]_j := \frac{\lambda_j}{\lambda_j + \lambda}[\hat{\beta}_0]_j,
\end{equation*}
where $\hat\beta_0$ is the ordinary least squares estimator.

Using the bias-variance decomposition, (Equation~\ref{biasDecomp}), we
have that:
\begin{lemma}
\begin{equation}
\nonumber
\mathrm{Risk} (\hat{\beta}_{\lambda}) = \frac{\sigma^2}{n}\sum_{j} \left(\frac{\lambda_j}{\lambda_j + \lambda} \right)^2 + \sum_{j} \beta^{2}_j \frac{\lambda_j}{(1 + \frac{\lambda_j}{\lambda})^2}.
\end{equation}
\end{lemma}
The proof is straightforward and is provided in the appendix.

\section{Ordinary Least Squares with PCA (PCA-OLS)}

Now let us construct a simple estimator based on $\lambda$. Note that
our rotated coordinate system where ${\bf \Sigma}$ is equal to $
diag(\lambda_1,\lambda_2,\ldots,\lambda_p)$ corresponds the PCA
coordinate system.

Consider the following ordinary least squares estimator on the ``top''
PCA subspace --- it uses the least squares estimate on coordinate $j$
if $\lambda_j\geq\lambda$ and $0$ otherwise
\begin{equation*}
[\hat{\beta}_{PCA,\lambda}]_j = \left\{ \begin{array}{rl}
  [\hat{\beta}_0]_j&\mbox{ if ${\lambda}_j \geq \lambda$} \\
  0 &\mbox{ otherwise}
       \end{array} \right.
.\end{equation*}
The following claim shows this estimator compares favorably to the
ridge estimator (for every $\lambda$)-- no matter how the $\lambda$ is chosen e.g., using cross validation or any other strategy.

\paragraph\\
Our main theorem (Theorem 2) bounds the Risk Ratio/Risk Inflation\footnote{Risk Inflation has also been used as a criterion for evaluating feature selection procedures~\citep{ric}.} of the PCA-OLS and the RR estimators.  

\begin{theorem}
(Bounded Risk Inflation) For all $\lambda\geq 0$, we have that:
\begin{equation}
\nonumber
0 \leq \frac{\mathrm{Risk}(\hat{\beta}_{PCA,\lambda})}{\mathrm{Risk}(\hat{\beta}_{\lambda})} \leq 4,
\end{equation}
and the left hand inequality is tight.
\end{theorem}

\begin{proof}
Using the bias variance decomposition of the risk we can write the risk as:
\begin{equation}
\nonumber
\mathrm{Risk}(\hat{\beta}_{PCA,\lambda}) = \frac{\sigma^2}{n}\sum_j
\mathbbm{1}_{\lambda_j \ge \lambda} + \sum_{j: \lambda_j < \lambda}
\lambda_j \beta_j^{2}.
\end{equation}
The first term represents the variance and the second the bias.

The ridge regression risk is given by Lemma 1.  We now show that the
$j^{th}$ term in the expression for the PCA risk is within a factor $4$
of the $j^{th}$ term of the ridge regression risk. First, let's consider
the case when $\lambda_j \ge \lambda$, then the ratio of $j^{th}$ terms
is: 
\[
\frac{\frac{\sigma^2}{n}}{\frac{\sigma^2}{n} \left(\frac{\lambda_j}{\lambda_j + \lambda} \right)^2 +  \beta_j^{2} \frac{\lambda_j}{(1 + \frac{\lambda_j}{\lambda})^2}} \leq \frac{\frac{\sigma^2}{n}}{\frac{\sigma^2}{n} \left(\frac{\lambda_j}{\lambda_j + \lambda} \right)^2}
= \left(1 + \frac{\lambda}{\lambda_j} \right)^2
\leq 4.
\]
Similarly, if $\lambda_j < \lambda$, the ratio of the $j^{th}$ terms is:
\[
\frac{\lambda_j \beta_j^{2}}{\frac{\sigma^2}{n} \left(\frac{\lambda_j}{\lambda_j + \lambda} \right)^2 +  \beta_j^{2} \frac{\lambda_j}{(1 + \frac{\lambda_j}{\lambda})^2}} \leq \frac{\lambda_j\beta_j^{2}}{\frac{\lambda_j\beta_j^{2}}{(1+ \frac{\lambda_j}{\lambda})^2}}
= \left(1+ \frac{\lambda_j}{\lambda}  \right)^2
\leq 4.
\]
Since, each term is within a factor of $4$ the proof is complete.
 \end{proof}
 
 It is worth noting that the converse is not true and the ridge regression estimator (RR) can be arbitrarily worse than the PCA-OLS estimator. An example which shows that the left hand inequality is tight is given in the Appendix.

\section{Experiments}
First, we generated synthetic data with $p=100$ and varying values of $n$= \{20, 50, 80, 110\}. The data was generated in a fixed design setting as   $Y = {\bf X}\beta + \epsilon$ where $\epsilon_i~\sim~\mathcal{N}(0,1)\;\;\;\forall i~=~1,\ldots, n$. Furthermore, ${\bf X}_{n\times p} \sim MVN({\bf 0}, {\bf I})$ where MVN(${\bf \mu}, {\bf \Sigma}$) is the Multivariate Normal Distribution with mean vector {$\bf \mu$}, variance-covariance matrix {$\bf \Sigma$} and $\beta_j \sim \mathcal{N}(0,1) \;\;\;\forall j~=~1,\ldots, p$.

\begin{figure*}[htbp]
\begin{center}

  \begin{tabular}{cccc}\hspace{-6mm}
    \begin{minipage}{4cm}
      \center{\epsfxsize=4cm
      \epsffile{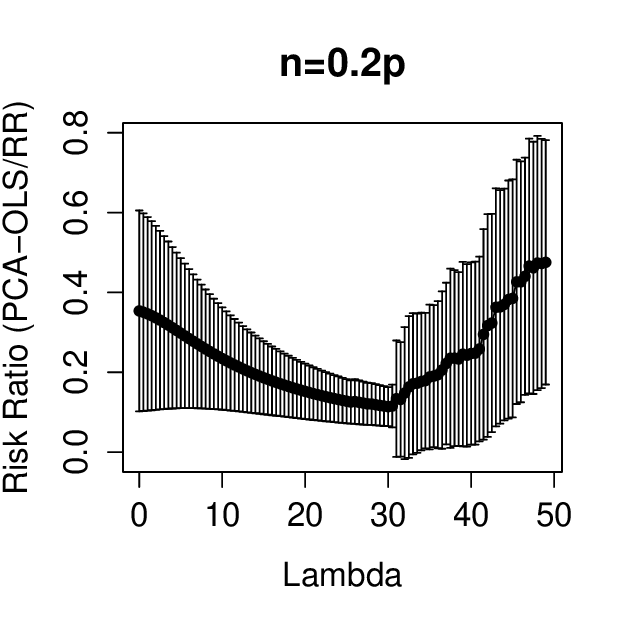}}\vspace{-2mm}
     \end{minipage}\hspace{-3mm}
    \begin{minipage}{4cm}
      \center{\epsfxsize=4cm
     \epsffile{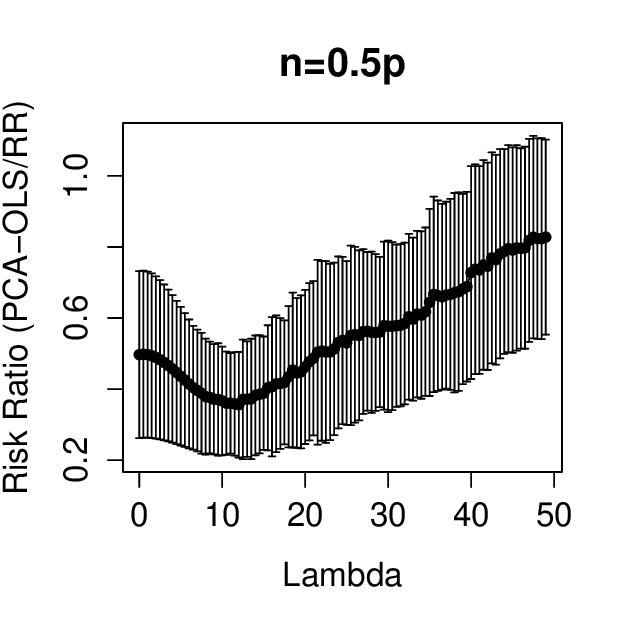}}\vspace{-2mm}
    \end{minipage}\hspace{-3mm}
     \begin{minipage}{4cm}
      \center{\epsfxsize=4cm
     \epsffile{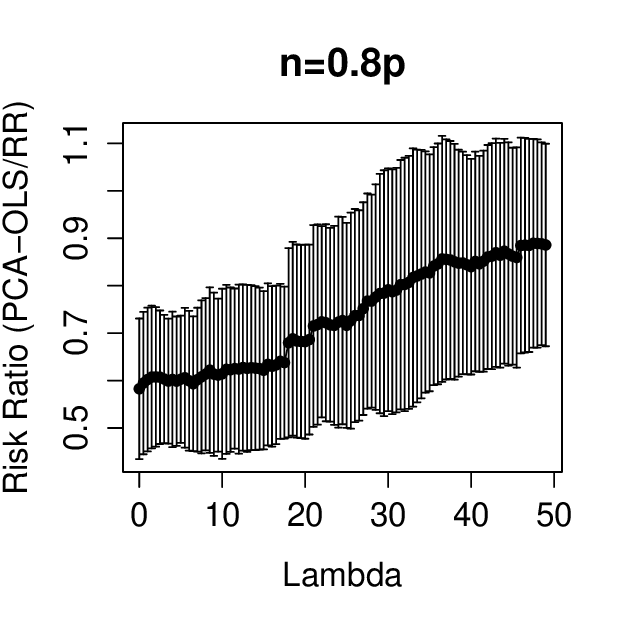}}\vspace{-2mm}
    \end{minipage}\hspace{-3mm}
     \begin{minipage}{4cm}
      \center{\epsfxsize=4cm
     \epsffile{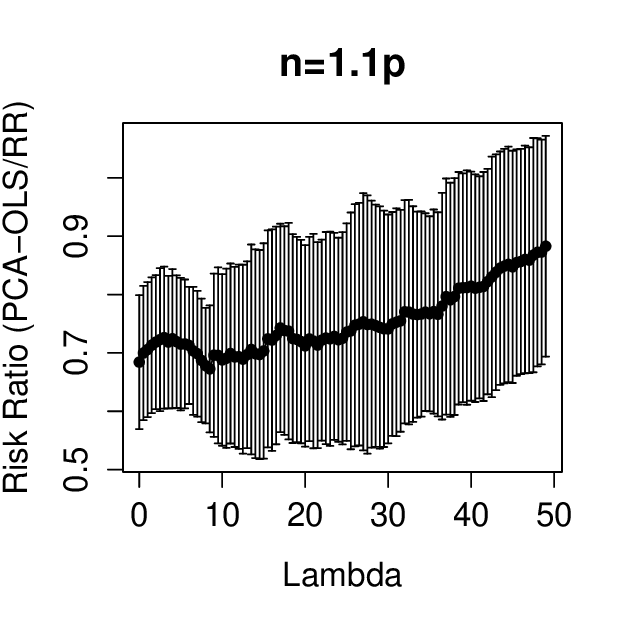}}\vspace{-2mm}
    \end{minipage}\hspace{-3mm}
        \vspace{-2mm}
  \end{tabular}

\vspace{.1in}
\caption{Plots showing the risk ratio as a function of $\lambda$, the regularization parameter and $n$, for the synthetic dataset. p=100 in all the cases.  The error bars correspond to one standard deviation for 100 such random trials.}
\label{tab:1}
\end{center}
\begin{center}

  \begin{tabular}{cccc}\hspace{-6mm}
    \begin{minipage}{4cm}
      \center{\epsfxsize=4cm
      \epsffile{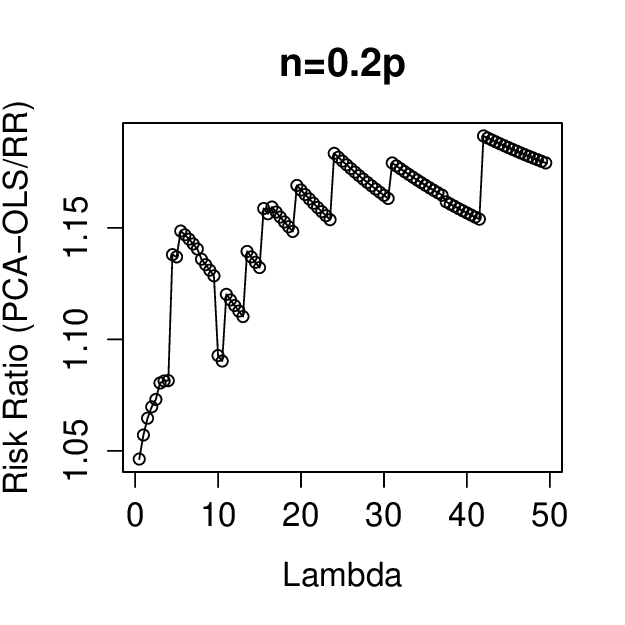}}\vspace{-2mm}
     \end{minipage}\hspace{-3mm}
    \begin{minipage}{4cm}
      \center{\epsfxsize=4cm
     \epsffile{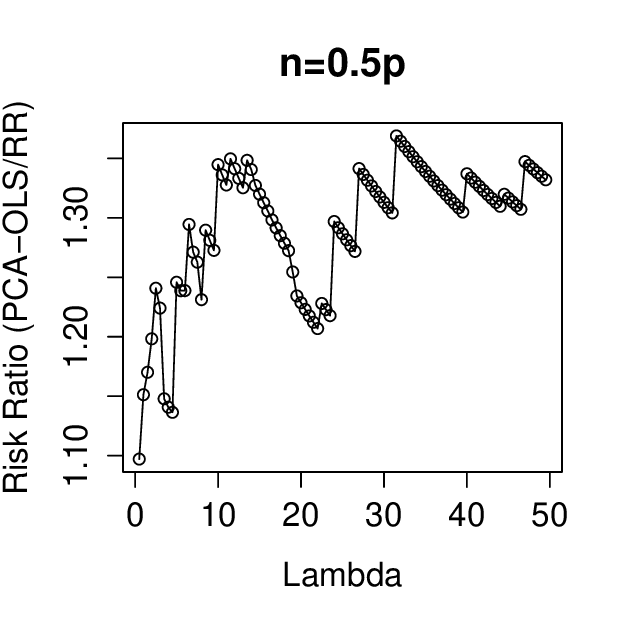}}\vspace{-2mm}
    \end{minipage}\hspace{-3mm}
     \begin{minipage}{4cm}
      \center{\epsfxsize=4cm
     \epsffile{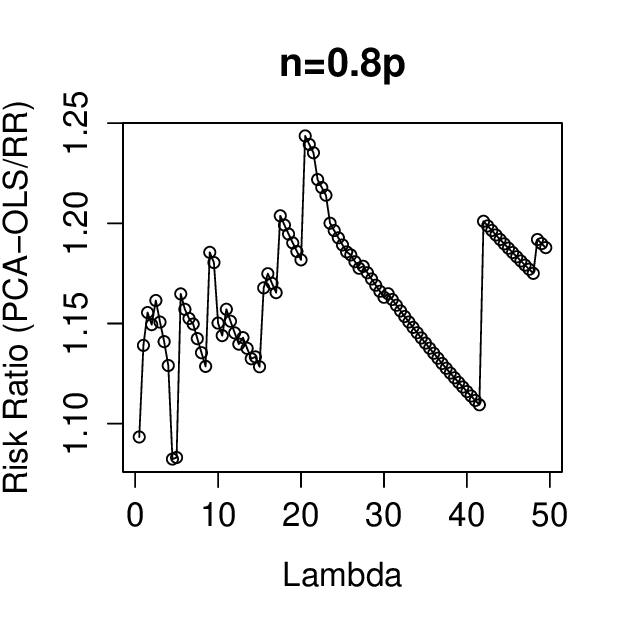}}\vspace{-2mm}
    \end{minipage}\hspace{-3mm}
     \begin{minipage}{4cm}
      \center{\epsfxsize=4cm
     \epsffile{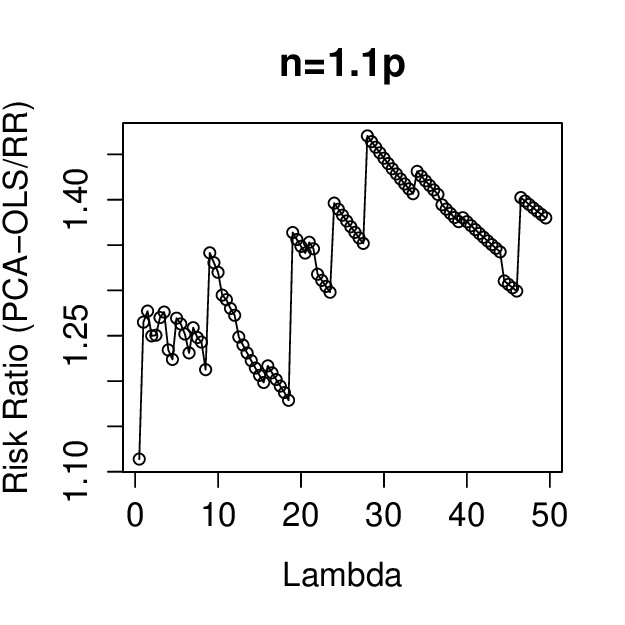}}\vspace{-2mm}
    \end{minipage}\hspace{-3mm}
  \end{tabular}
  
  \begin{tabular}{cccc}\hspace{-6mm}
    \begin{minipage}{4cm}
      \center{\epsfxsize=4cm
      \epsffile{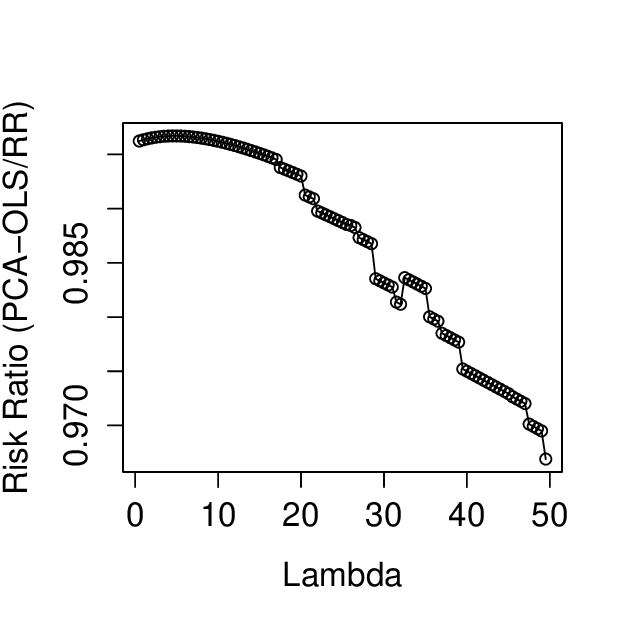}}\vspace{-2mm}
     \end{minipage}\hspace{-3mm}
    \begin{minipage}{4cm}
      \center{\epsfxsize=4cm
     \epsffile{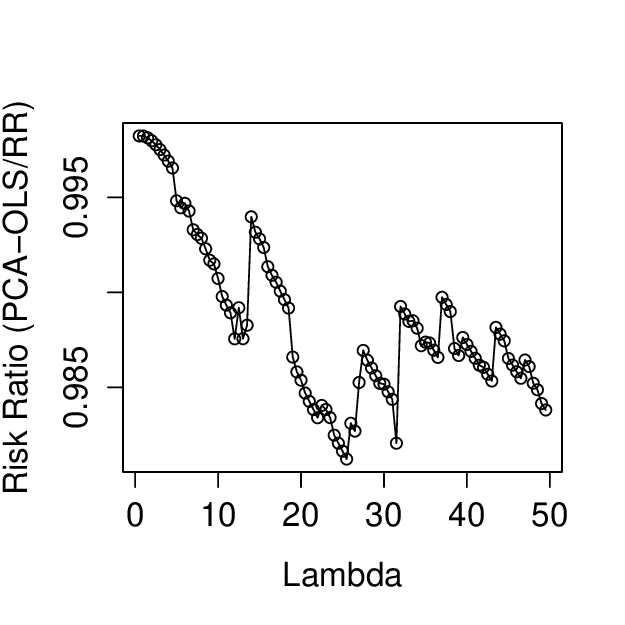}}\vspace{-2mm}
    \end{minipage}\hspace{-3mm}
     \begin{minipage}{4cm}
      \center{\epsfxsize=4cm
     \epsffile{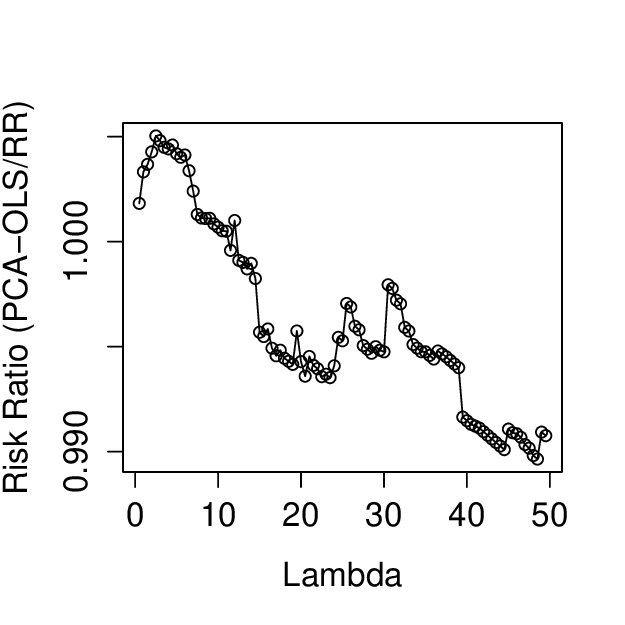}}\vspace{-2mm}
    \end{minipage}\hspace{-3mm}
     \begin{minipage}{4cm}
      \center{\epsfxsize=4cm
     \epsffile{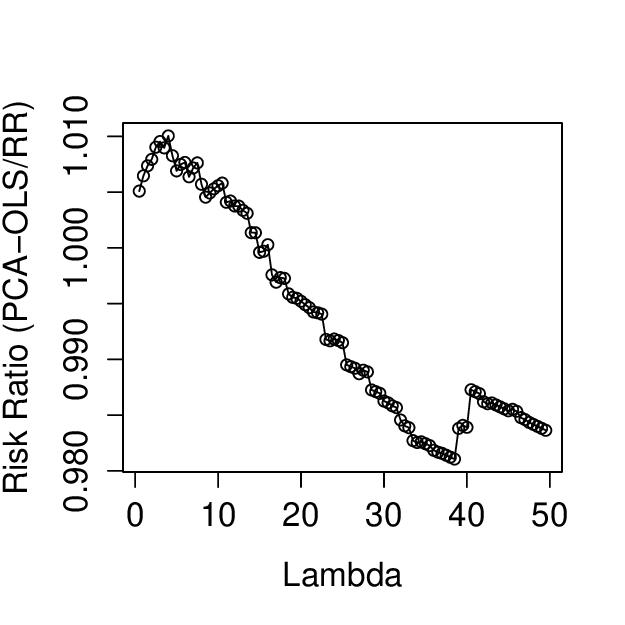}}\vspace{-2mm}
    \end{minipage}\hspace{-3mm}
        \vspace{-2mm}
  \end{tabular}

\caption{Plots showing the risk ratio as a function of $\lambda$, the regularization parameter and $n$, for two real world datasets (BCI and USPS--top to bottom). }
\label{tab:2}
\end{center}
\end{figure*}

The results are shown in Figure~\ref{tab:1}. As can be seen, the risk ratio of PCA (PCA-OLS) and ridge regression (RR) is never worse than 4 and often its better than 1 as dictated by Theorem 2.

Next , we chose two real world datasets, namely USPS (n=1500, p=241) and BCI (n=400, p=117)\footnote{The details about the datasets can be found here:  \url{http://olivier.chapelle.cc/ssl-book/benchmarks.html.}}. 

Since we do not know the true model for these datasets,  we used all the $n$ observations to fit an OLS regression and used it as an estimate of the true parameter $\beta$. This is a reasonable approximation to the true parameter as we estimate the ridge regression (RR) and PCA-OLS models on a small subset of these observations. Next we choose a random subset of the observations, namely $0.2\times p$, $0.5\times p$ and $0.8\times p$ to fit the ridge regression (RR) and PCA-OLS models.

The results are shown in Figure~\ref{tab:2}. As can be seen, the risk ratio of PCA-OLS to ridge regression (RR) is again within a factor of 4 and often PCA-OLS is better i.e., the ratio $<1$.

\section{Conclusion}

We showed that the risk inflation of  a particular ordinary least
squares estimator (on the ``top'' PCA subspace) is within a
factor 4 of the ridge estimator. 
It turns out the converse is not true
--- this PCA estimator may be arbitrarily better than the ridge one.

\appendix
\section{}

\noindent{\bf Proof of Lemma 1.}
\begin{proof}
We analyze the bias-variance decomposition in
Equation~\ref{biasDecomp}. For the variance,
\begin{eqnarray}
\nonumber
\mathbb{E}_Y\|\hat{\beta}_{\lambda}-\bar{\beta}_{\lambda}\|^2_{\bf \Sigma}&=& \sum_j \lambda_j\mathbb{E}_Y([\hat{\beta}_{\lambda}]_j- [\bar{\beta}_{\lambda}]_j)^2\\
\nonumber
&=& \sum_j \frac{\lambda_j}{(\lambda_j + \lambda)^2}\frac{1}{n^2}\mathbb{E}\left[\sum_{i=1}^n(Y_i - \mathbb{E}[Y_i])[X_i]_j\sum_{i'=1}^n(Y_i' - \mathbb{E}[Y_i'])[X_i']_j\right]\\
\nonumber
&=&\sum_j \frac{\lambda_j}{(\lambda_j + \lambda)^2}\frac{\sigma^2}{n}\sum_{i=1}^n Var(Y_i)[X_i]^2_j\\
\nonumber
&=& \sum_j \frac{\lambda_j}{(\lambda_j + \lambda)^2} \frac{\sigma^2}{n} \sum_{i=1}^n[X_i]_j^2\\
&=& \frac{\sigma^2}{n}\sum_j \frac{\lambda_j^2}{(\lambda_j +\lambda)^2}.
\nonumber
\end{eqnarray}
 Similarly, for the bias,
\begin{eqnarray}
\nonumber
\|\bar{\beta}_{\lambda}-\beta\|^2_{\bf \Sigma} &=& \sum_j \lambda_j ([\bar{\beta}_{\lambda}]_j- [\beta]_j)^2\\
\nonumber
&=& \sum_j \beta_j^{2} \lambda_j \left(\frac{\lambda_j}{\lambda_j + \lambda}-1 \right)^2\\
&=& \sum_j \beta_j^{2}  \frac{\lambda_j}{(1+\frac{\lambda_j}{\lambda})^2},
\nonumber
\end{eqnarray}
which completes the proof.
\end{proof}

\noindent{\bf The risk for RR can be arbitrarily worse than the PCA-OLS estimator.}

Consider the standard OLS setting described in Section 1 in which ${\bf X}$ is $n\times p$ matrix and $Y$ is a $n \times 1$ vector. 

Let ${\bf X}=diag(\sqrt{1+\alpha},1,\ldots,1)$,  then ${\bf \Sigma}= {\bf X}^{\top}{\bf X}= diag(1+\alpha,1,\ldots,1)$ for some ($\alpha >0$) and also choose $\beta=[2+\alpha,0,\ldots,0]$.
For convenience let's also choose $\sigma^2=n$.

Then, using Lemma 1, we get the risk of RR estimator as

\begin{equation}
\nonumber
\mathrm{Risk} (\hat{\beta}_{\lambda}) = \left(\underbrace{\left(\frac{1+\alpha}{1+\alpha+\lambda}\right)^2}_{\text{I}} + \underbrace{\frac{(p-1)}{(1+\lambda)^2}}_{\text{II}}\right)  + \underbrace{(2+\alpha)^2\times \frac{(1+\alpha)}{(1+ \frac{1+\alpha}{\lambda})^2}}_{\text{III}}.
\end{equation}

Let's consider two cases

\begin{itemize}
\item {\bf Case 1:} $\lambda < (p-1)^{1/3}-1$, then $II > (p-1)^{1/3}$.
\item {\bf Case 2:} $\lambda > 1$, then $1+ \frac{1+\alpha}{\lambda} < 2+\alpha$, hence $III > (1+\alpha)$. 
\end{itemize}

Combining these two cases we get $\forall \lambda$, $\mathrm{Risk} (\hat{\beta}_{\lambda}) > min((p-1)^{1/3}, (1+\alpha))$. If we choose $p$ such that $p-1= (1+\alpha)^3$, then $\mathrm{Risk} (\hat{\beta}_{\lambda}) > (1+\alpha)$.

The PCA-OLS risk (From Theorem 2) is: 

\begin{equation}
\nonumber
\mathrm{Risk}(\hat{\beta}_{PCA,\lambda}) = \sum_j
\mathbbm{1}_{\lambda_j \ge \lambda} + \sum_{j: \lambda_j < \lambda}
\lambda_j \beta_j^{2}.
\end{equation}

Considering $\lambda \in (1, 1+\alpha$), the first term will contribute 1 to the risk and rest everything will be 0. So the risk of PCA-OLS is $1$ and the risk ratio is 

\begin{equation}
\nonumber
\frac{\mathrm{Risk}(\hat{\beta}_{PCA,\lambda})}{\mathrm{Risk} (\hat{\beta}_{\lambda}) }\leq \frac{1}{(1+\alpha)}.
\end{equation}

Now, for large $\alpha$, the risk ratio $\approx 0$.

\bibliography{keeporkill_risk}

\end{document}